\documentclass[a4paper,twoside]{article}

\usepackage{epsfig}
\usepackage{subcaption}
\usepackage{calc}
\usepackage{amssymb}
\usepackage{amstext}
\usepackage{amsmath}
\usepackage{amsthm}
\usepackage{multicol,multirow}
\usepackage{pslatex}
\usepackage{apalike}
\usepackage{algorithm2e}
\usepackage[bottom]{footmisc}
\usepackage{algorithmic}

\usepackage{hyperref}

\usepackage[usenames,dvipsnames]{xcolor}
\usepackage[breakable, theorems, skins]{tcolorbox}

\usepackage{SCITEPRESS}     

\begin{document}

\title{Improving Implicit Hate Speech Detection via a Community-Driven Multi-Agent Framework}

\author{\authorname{Ewelina Gajewska\sup{1}\orcidAuthor{0009-0006-6012-4787}, Katarzyna Budzynska\sup{1}\orcidAuthor{0000-0001-9674-9902} and Jarosław A. Chudziak\sup{1}\orcidAuthor{0000-0003-4534-8652}}
\affiliation{\sup{1}Warsaw University of Technology, Plac Politechniki 1, Warsaw, Poland}
\email{ewelina.gajewska.dokt@pw.edu.pl, katarzyna.budzynska@pw.edu.pl, jaroslaw.chudziak@pw.edu.pl}
}


\abstract{This work proposes a contextualised detection framework for implicitly hateful speech, implemented as a multi-agent system comprising a central Moderator Agent and dynamically constructed Community Agents representing specific demographic groups. Our approach explicitly integrates socio-cultural context from publicly available knowledge sources, enabling identity-aware moderation that surpasses state-of-the-art prompting methods (zero-shot prompting, few-shot prompting, chain-of-thought prompting) and alternative approaches on a challenging ToxiGen dataset. We enhance the technical rigour of performance evaluation by incorporating balanced accuracy as a central metric of classification fairness that accounts for the trade-off between true positive and true negative rates. We demonstrate that our community-driven consultative framework significantly improves both classification accuracy and fairness across all target groups. 
}

\onecolumn \maketitle \normalsize \setcounter{footnote}{0} \vfill


\section{\uppercase{Introduction}}
\label{sec:introduction}

The widespread proliferation of harmful online content poses a substantial threat to inclusive digital environments and user well-being. 
The vast and continually growing volume of user-generated content on social media platforms presents a substantial challenge for content moderation, necessitating the development and deployment of automated information processing technologies \cite{gillespie2020content,huang2025content}. To address this scale, platforms increasingly rely on advanced machine learning techniques, fine-tuning of compact Transformer-based models, and most recently, prompting strategies applied to large language models (LLMs) using custom datasets. 

While these techniques contribute to progressive improvements in overall classification accuracy, a critical concern remains regarding their fairness performance, specifically, the balance between true negative and true positive rates \cite{kolla2024llm,thiago2021fighting}. Most current systems demonstrate high true negative rates, effectively identifying non-hateful content, yet suffer from comparatively low true positive rates, thereby under-detecting hateful content. This imbalance hampers equitable moderation outcomes, disproportionately affecting marginalised communities and highlighting the need for more nuanced, context-aware approaches to hate speech detection. 

This limitation is compounded by the fact that hate speech interpretation critically depends on the identities involved. It is due to the fact that hate speech is not merely a linguistic phenomenon but an intrinsically social and contextual construct; thus, it can be negotiated between individuals and groups \cite{sap2021annotators}. Current automated systems commonly embody dominant viewpoints embedded in their training data, failing to systematically integrate minority perspectives. Consequently, these systems often misclassify potentially hateful content, undermining trust in moderation efforts and exacerbating social inequalities \cite{vaccaro2021contestability}. 
A solution to this problem involves incorporating some form of expert knowledge into the moderation process, enabling context-sensitive and identity-aware evaluations that better reflect the lived experiences of targeted communities.

This work aims to address these critical research insights, offering a practical solution that operationalises fairness principles to improve the effectiveness and equity of content moderation decisions in online environments. 
Specifically, we propose a multi-agent moderation system, which integrates a central Moderator Agent with specialised Community Agents representing the contextualised information about targeted groups, following findings on the role of expertise in content moderation \cite{abdelkadir2025role}. It enables adaptive, consultative classification, where the moderator solicits community expertise on ambiguous cases, thereby enhancing contextual sensitivity of this decision-making process. 

Our research is guided by two primary research questions: 
(1) How does integrating community-driven consultation influence the moderation system’s final classification decisions, especially in ambiguous or coded language scenarios? 
(2) Does this consultative framework improve classification fairness more than standard moderation approaches? 
Through experimental testing, we show that a community-driven design underpinning our content moderation system yields significant benefits for both detection accuracy and moderation fairness. 
This work makes the following contributions: 
(1) We present a multi-agent community-driven detection framework for implicitly hateful speech. 
(2) We validate the proposed architecture on a demographically diverse dataset, demonstrating substantial improvements in both classification accuracy and fairness relative to established prompting-based benchmarks. 
This establishes our framework as a practical advance in agent-based artificial intelligence for content moderation tasks.


\section{\uppercase{Related Work}}
The introduction of Transformer-based models has brought notable improvements in identifying implicit forms of hate speech by better encoding contextual information within texts \cite{yadav2024hatefusion,nghiem2024define}. 
These models outperformed earlier architectures by grasping subtle cues and indirectly hateful terms embedded in context. Nonetheless, their effectiveness hinges predominantly on a fine-tuning process and therefore on the availability of high-quality annotated datasets, which are costly and time-consuming to produce. As a result, data scarcity and annotation quality remain significant bottlenecks for Transformer-based hate speech classifiers, impeding their broad applicability and generalisability.

Recently, LLM-based prompting methods have been developed to enhance the effectiveness of this task. Several studies have explored zero-shot and few-shot prompting strategies, showing that properly framed queries can substantially improve model performance without fine-tuning process \cite{jaremko2025revisiting,pan2024comparing}. 
However, both strategies involve in-context learning, where the LLM leverages context provided in the prompt for task adaptation without gradient-based updates; therefore, the success of these approaches depends heavily on prompt design and the quality of examples in prompts \cite{garcia2023leveraging,plaza2023respectful}. 

To this end, Chain-of-Thought (CoT) and rationale-augmented prompting approaches have been proposed to encourage models to reason explicitly about intent, definitions, and moderation norms, leading to more reliable outcomes in ambiguous cases \cite{kojima2022large,vishwamitra2024moderating}. They do so by guiding LLMs to generate intermediate reasoning steps before arriving at a final classification, effectively decomposing complex decisions into a sequence of manageable cognitive operations \cite{kojima2022large}. 
Further enhancements are realised by Decision-Tree-of-Thought (DToT) prompting with an iterative mechanism that selectively re-prompts LLMs with increasingly fine-grained context \cite{zhang2024efficient}. 
Despite these innovations, empirical results demonstrate that DToT only marginally surpasses traditional CoT, indicating diminishing returns for increased complexity \cite{zhang2024efficient}. 

Limitations of commercial moderation APIs further expose these challenges as recent analysis reveals that many commercial systems tend to over-rely on surface-level group identity markers, resulting in both under-moderation of implicit hate speech and over-moderation of legitimate speech acts like counter-speech or reclaimed language  \cite{wang2024human}. Furthermore, LLM moderators fail when faced with ambiguity or subtlety in meaning, reinforcing the essential role of human judgment \cite{kolla2024llm}. 

The picture is clear: current approaches fundamentally depend on text-centric and model-centric enhancements, lacking explicit mechanisms for integrating the rich social and cultural contexts that shape implicit hate speech. Sociocultural research emphasises that hate speech is deeply embedded in cultural norms and societal values, influencing how harmful content is perceived and understood \cite{aoyagui2025matter,plum2025identity,udupa2023ethical}. For instance, \cite{aoyagui2025matter} explored explanation styles of humans and LLMs in subjective moderation tasks related to subtle sexism, revealing that while both groups manifest consistent perspectival framings, their distributions and interpretive nuances differ. Therefore, without accounting for these socio-cultural dimensions, existing methods risk misclassifying or overlooking covert hateful intentions.

\begin{figure*}[ht!]
    \centering
    \includegraphics[width=0.95\linewidth]{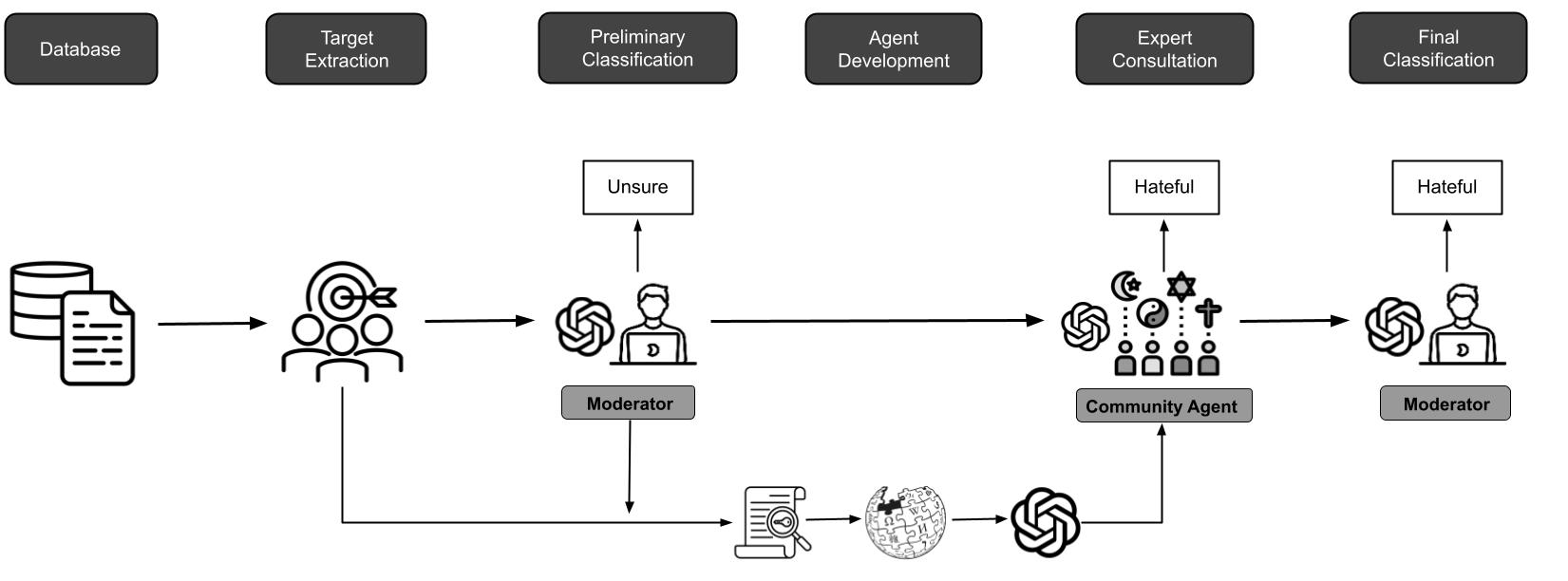}
    \caption{The design of our multi-agent consultative system.}
    \label{fig:systemdesign}
\end{figure*}

To this end, social computing research has explored the role of persona-agents and generative agents that simulate human behaviour and social interactions in virtual environments \cite{gajewska2025algorithmic,park2023generative,choi2025proxona,zamojska2025games}. These agents enable the creation of populated and interactive systems that mimic human decision-making and social behaviours. However, despite promising applications in domains such as creative ideation, there exists limited empirical study on the utility of persona-agents for content moderation tasks. In particular, questions remain about whether and how such agents can effectively embody diverse social identities and perspectives to improve moderation fairness and contextual sensitivity. 
Our work addresses this gap by introducing a multi-agent framework in which Community Agents instantiate identity-aware expert perspectives. The effectiveness of this approach is empirically evaluated with respect to moderation accuracy and fairness, with the aim of advancing more adaptive and socially sensitive moderation practices.

\section{\uppercase{Methodology}}
In this section, the methodological design of the proposed moderation framework is detailed. A simulation-based approach is adopted to model the content moderation process within a multi-LLM environment, in which LLM-based community agents are integrated into a dynamic, consultative workflow aimed at enhancing fairness and contextual sensitivity of hate speech detection on social media platforms.








\subsection{System Architecture}
The workflow of our moderation system is depicted in Figure \ref{fig:systemdesign}. At the core of this framework is the Moderator Agent, tasked with the initial semantic and contextual evaluation of social media posts. Operating according to robust moderation policies, the Moderator closely emulates the consistency and rule compliance expected from trained human moderators. 
In contrast, Community Agents serve as expert advisers, each embodying the unique perspective of a particular socio-cultural group targeted by hate. 
This approach resonates with established principles of contestability and expert-in-the-loop governance, ensuring that system output can be challenged, corrected, and iteratively improved with direct engagement from embedded domain specialisation \cite{vaccaro2021contestability}. 
By capturing the expert knowledge of marginalised communities in agent design, our architecture operationalises participatory involvement and reconfigures automated moderation toward greater fairness and domain relevance. 

\paragraph{Community Agent Construction.} 
To simulate identity-aware perspectives in our moderation system, we employ a multi-stage process for agent personality creation that leverages LLMs’ contextual and generative capacities. The identification of the target group, as supplied by metadata or preliminary automated classification, triggers the automatic generation of multiple search queries tailored to the community in question. These queries retrieve background information from the Wikipedia knowledge base, anchoring agent construction in public, factual data to minimise the risk of hallucinated or biased outputs. 
For instance, when evaluating a text that advocates for America ``a return to how things were in the South before'', Wikipedia provides rich detail on the period of slavery, the Jim Crow laws, and the systematic segregation inflicted on African Americans throughout the 19th and 20th centuries. This contextual information is indispensable for correctly interpreting coded language and historical allusions used in implicit hate speech, such as calls for regression to a racially discriminatory legal order.

To operationalise these contextually grounded agent profiles within the moderation system, the retrieved and synthesised information must be transformed into a structured representation suitable for computational reasoning. This process involves encoding the collected textual data into embeddings, which serve as the foundation for downstream moderation decisions. 
Let \(W_g=\{w_1, w_2, \dots, w_{K_d}\}\) denote the set of retrieved Wikipedia texts related to the detected group \(g\). 
Each document \(w_i\) is tokenized and encoded by a Transformer encoder into contextual token embeddings: 
\[
\mathbf{H}_i = [\mathbf{h}_{i1},\dots,\mathbf{h}_{iT_i}] \in \mathbb{R}^{T_i \times d_h},
\]
where \(\mathbf{h}_{it}\) denotes the hidden embedding of the \(t\)-th token of document \(i\) and \(d_h\) is the hidden dimensionality.
Concatenating all token embeddings from the retrieved documents yields: 
\[
\mathbf{H} = [\mathbf{H}_1,\dots,\mathbf{H}_{K_d}] \in \mathbb{R}^{T_{\mathrm{tot}} \times d_h},\qquad
T_{\mathrm{tot}}=\sum_{i=1}^{K_d} T_i.
\] 
We obtain the group (query) embedding \(\mathbf{q}_g=\phi(g)\), and define the community agent embedding \(\psi_g\) via cross-attention between \(\mathbf{q}_g\) and the retrieved token embeddings:
\[
\psi_g = \mathrm{Attention}\!\big(\mathbf{q}_g,\; \mathbf{K},\; \mathbf{V}\big),
\]
where \(\mathbf{Q}=q_g,\mathbf{K}=\mathbf{V}=\mathbf{H}\), and the standard scaled dot-product attention is
\[
\mathrm{Attention}(Q,K,V)=\mathrm{softmax}\!\left(\frac{QK^\top}{\sqrt{d_k}}\right) V.
\]

\subsection{Multi-Agent Consultative Workflow} 
The moderation process unfolds in a structured three-phase procedure: preliminary assessment, conditional expert consultation, and final decision synthesis. 
The Moderator serves as the central decision-making agent, tasked with an initial classification of social media posts. Upon receiving a post and the associated target group, the Moderator conducts a preliminary analysis to determine whether the content constitutes hate speech. This preliminary step culminates in three outputs: an initial classification label (Hate Speech / Not Hate Speech / Unsure), a justification of this decision, and a meta-cognitive flag indicating whether the Moderator requires additional insights by consulting a Community Agent (Consultation Needed: Yes / No). 
Crucially, Community Agents are activated only when explicitly requested by the Moderator Agent via a targeted consultation query. 
Finally, the Moderator synthesises Community Agent’s input with its initial assessment to produce a final, informed classification decision. This decision includes an integrated justification explicitly acknowledging how community-driven insights affected the outcome. If no consultation was triggered, the initial evaluation stands as conclusive. 

Research consistently shows that content moderation is a subjective, often contentious task influenced by moderators’ backgrounds, beliefs, as well as interpretations of context and intent \cite{sap2021annotators,yoder2022hate}. Ambiguity arises from subtle, implicit, or coded language where the boundary between hate speech and non-hate speech can be unclear, leading to disagreements even among expert moderators \cite{sang2022origin}. As a result, scholars highlight the need for domain expertise when reviewing social media data from vulnerable populations \cite{patton2019annotating}. 
Our methodology operationalises this concept of identity-dependent perception, allowing the moderator to seek expert in-group validation on ambiguous cases, thereby aiming to produce more accurate and fairer content moderation outcomes. Fair content moderation entails the provision of equitable treatment across diverse target groups, with particular emphasis on the accurate identification and protection of marginalised populations, thereby mitigating risks of under-detection, misclassification, and disproportionate harm within these communities. 

\begin{algorithm}
\caption{Consultative multi-agent content moderation process}
\begin{algorithmic}[1]

\STATE \textbf{Inputs:} 
\STATE \quad $x$ \COMMENT{input text sample}
\STATE \quad $\tau_{low}, \tau_{high}$ \COMMENT{moderator uncertainty bounds}
\STATE \quad $K_d = 5$ \COMMENT{number of Wikipedia queries} 

\STATE \textbf{Outputs:} 
\STATE \quad $p_{final}$ \COMMENT{final decision score}


\STATE \textbf{Procedure:}

\STATE \textbf{1. Target Extraction}
\STATE $g \gets ExtractTargetGroup(x)$


\STATE \textbf{2. Initial Evaluation}
\STATE $(p_m, r_m) \gets ModeratorAgent(x, g)$
\IF{$p_m \in [\tau_{low}, \tau_{high}]$}
    \STATE $trigger\_persona \gets True$
\ELSE
    \STATE $trigger\_persona \gets False$
\ENDIF


\STATE \textbf{3. Persona Construction and Decision Fusion}
\IF{$trigger\_persona$}
    \STATE $Q \gets GenerateWikipediaQueries(g, K_d)$
    \STATE $W_g \gets RetrieveWikipediaPage(q_i) \mid i=1..K_d$
    \STATE $\psi_g \gets BuildCommunityPersona(W_g)$ 
    
    \STATE $(p_c, r_c) \gets CommunityAgent(x, \psi_g)$
    \STATE $p_{final} \gets CombineScores(p_m, p_c, r_m, r_c)$    
\ELSE
    \STATE $p_{final} \gets p_m$
\ENDIF


\STATE \textbf{4. Return}
\STATE \quad $p_{final}$
\end{algorithmic}
\end{algorithm}

The entire framework operates within AutoGen library, leveraging generative LLMs (Gemini-2.5-Flash) for both the Moderator Agent and Community Agents. Each agent is carefully configured with a distinct system prompt that dictates its behavioural heuristics, response format, and interaction protocols. This structured setup ensures reproducibility and clarity in agent communication, mimicking real-world human moderator workflows and consultations.

\subsection{Datasets and Evaluation}
For system validation, we employ a dataset that includes a range of implicit hate speech examples targeting multiple demographic groups - Toxigen \cite{hartvigsen2022toxigen}. It comprises approximately 274,000 short texts (with 100 characters on average representing social media contexts), balanced between toxic and benign examples, of which 8,960 have been manually annotated by Mechanical Turk workers. This demographic breadth aligns closely with our system’s objective to integrate identity-aware moderation via specialised Community Agents. Another defining characteristic of ToxiGen, which challenges automatic moderation systems, is its emphasis on implicit toxicity, with over 98\% of toxic examples featuring subtle, context-dependent expressions rather than explicit hate speech. This property makes ToxiGen particularly well-suited for studying challenging, culturally grounded forms of hate speech that conventional moderation protocols often overlook, directly supporting our approach of consultative community-driven moderation to enhance contextual accuracy and ethical alignment. For the purpose of our study we choose two racial groups (Black people, Asians), two religious groups (Muslims, Jewish people) and two gender-related groups (Women, LGBTQ) and therefore generate six Community Agents in addition to the central Moderator. 

For testing purposes, we utilise the manually annotated part of the Toxigen dataset and randomly sample 100 cases for each target group. 
Evaluation metrics focus on key performance indicators that assess both the effectiveness and fairness of hate speech detection. The F1 score serves as the metric of classification effectiveness. It provides a balanced measure of a classifier’s precision and recall rates. 
Then, True Positive Rate (TPR) quantifies the proportion of actual hate speech instances correctly identified by the system. A high TPR is critical to ensuring that harmful content is effectively flagged, thereby protecting marginalised groups from digital harm. Conversely, the True Negative Rate (TNR) measures the proportion of benign content that is correctly classified as such, reflecting the system’s ability to avoid over-moderation, which can suppress legitimate expression. These metrics are paramount in the context of existing research with human and automated moderation systems, which frequently report high TNR paired with low TPR, indicating a conservative bias that favours avoiding false positives at the cost of under-detecting hate speech \cite{kolla2024llm}. 
To unify these two metrics, we employ Balanced Accuracy (bACC) as a primary metric of classification fairness. It equally weights the TPR and TNR across classes, providing a single, interpretable value reflecting both sensitivity and specificity and mitigating bias that can arise from class imbalance. 
  
The evaluation framework is designed to assess the efficacy of the proposed multi-agent moderation system under a range of operational configurations. First, our agentic approach is contrasted with a baseline methodology employing CoT prompting, zero-shot prompting (Z-S) and few-shot (F-S) prompting with the same underlying LLM architecture (Gemini 2.5), enabling a direct comparison of our agentic versus prompt-based paradigms under comparable model capacity. Additionally, an ablation study evaluates the system variant in which the central Moderator independently classifies content without consultation with Community Agents, isolating the contribution of agent collaboration and contestability. Finally, we discuss the classification performance of our system in regard to alternative systems validated on the ToxiGen dataset, situating our method relative to state-of-the-art approaches documented in literature. 


\section{\uppercase{Results}}

\begin{table*}[ht]
    \centering
    \caption{Evaluation of classification results across six target groups and on average. In each case, we list three baselines and our Agentic approach (marked in bold). We first show true positive (TPR) and true negative rates (TNR) as we address the issue of balancing the identification of hateful and benign statements. Then, balanced accuracy (bACC) and F1 scores are reported to provide complementary perspectives on the overall performance. Asterisks (*) indicate approaches with the highest performance in each case. Finally, in parentheses (()) we report results from the ablation study of our Agentic approach without the Community Agents consultation.} \label{tab:results}
    \begin{tabular}{|ll|ll|lll|}
    \hline %
    Target &  Approach  & TPR & TNR & bACC & F1 Overall & F1-hate \\      \hline 
    
    Black &  Z-S Prompting & 75.0 & 90.0 & 82.5 & 0.830 & 0.789 \\ 
      & F-S Prompting & 82.0 & 87.0 & 84.5 & 0.844 & 0.815\\
       &   CoT Prompting & 82.0 & 90.0  & 86.0 & 0.864  & 0.835 \\  
       &  \textbf{Agentic}  &   \textbf{80.0} (64.0) & \textbf{93.3} (100) & \textbf{86.7}* (82.0) & \textbf{0.873}* (0.867)  &  \textbf{0.842}* (0.778) \\      
       \hline
        
      Asian &  Z-S Prompting & 22.0 & 100 & 61.0 & 0.540 & 0.361 \\ 
        & F-S Prompting & 40.0 & 98.0 & 69.0 &  0.662 & 0.563\\
         &   CoT Prompting & 54.0 & 98.0  & 76.0 & 0.748  & 0.692 \\  
         & \textbf{Agentic}  &  \textbf{68.0} (39.0) & \textbf{96.0} (100)  & \textbf{82.0}* (69.5) & \textbf{0.816}* (0.720) &  \textbf{0.791}* (0.560) \\        
         \hline
        
      Jewish &   Z-S Prompting & 35.0 & 98.0 & 66.5 &  0.618  & 0.507 \\ 
      & F-S Prompting & 50.0 & 96.0 & 73.0 & 0.708 & 0.650 \\
       &   CoT Prompting & 46.0 & 94.0  & 70.0 & 0.676  & 0.608 \\ 
       & \textbf{Agentic}  &  \textbf{75.0} (47.0) & \textbf{93.8} (100) & \textbf{84.4}* (73.5) & \textbf{0.839}* (0.766) &  \textbf{0.830}* (0.643) \\      
       \hline
                
      Muslim  &  Z-S Prompting  & 33.0 & 98.0  & 65.5 & 0.624 &  0.492 \\ 
       & F-S Prompting & 52.0 & 98.0 & 75.0 &  0.743 & 0.676 \\
       &   CoT Prompting & 56.0 & 98.0  & 77.0 & 0.767  & 0.711 \\  
       & \textbf{Agentic}  &  \textbf{79.2} (33.0) & \textbf{98.1} (100) & \textbf{88.7}* (66.5) & \textbf{0.888}* (0.701) & \textbf{0.874}* (0.500) \\       
       \hline
                 
     Women &   Z-S Prompting & 45.0 & 100 & 72.5 & 0.700 & 0.622 \\ 
      & F-S Prompting & 43.0 & 100 & 71.5 & 0.687 & 0.603 \\
      &   CoT Prompting & 59.0 & 98.0  & 78.5 & 0.773 & 0.732 \\ 
       & \textbf{Agentic}  &   \textbf{70.6} (28.0) & \textbf{100} (100) & \textbf{85.3}* (64.0) & \textbf{0.847}* (0.659) & \textbf{0.828}* (0.435) \\     
      \hline
               
     LGBTQ &   Z-S Prompting & 43.0 & 100 & 71.5 & 0.710  & 0.603 \\ 
      & F-S Prompting & 41.0 & 100 & 70.5 & 0.696  & 0.581 \\
       &   CoT Prompting & 59.0 & 95.0  & 77.0 & 0.773 & 0.712 \\ 
    & \textbf{Agentic}  &  \textbf{77.2} (0.0) & \textbf{100} (100) & \textbf{88.6}* (50.0) &  \textbf{0.895}* (0.472) & \textbf{0.872}* (0.0) \\

         \hline \hline
       Average &  Z-S Prompting &  42.2 & 97.7 & 70.0 & 0.670 & 0.562 \\
      & F-S Prompting & 51.3 & 96.5 & 73.9 & 0.723 & 0.648 \\
       &   CoT Prompting & 59.3 & 95.5  & 77.4 & 0.767  & 0.715 \\  
         &  \textbf{Agentic}  &  \textbf{75.0} (35.2) & \textbf{96.9} (100) & \textbf{86.0}* (67.6) & \textbf{0.860}* (0.698) & \textbf{0.840}* (0.486) \\
                 
        \hline
        \multicolumn{7}{l}{\footnotesize Differences in performance between agentic and other approaches are significant at $p<0.001$ (tested with paired t-tests)}
    \end{tabular} 
    
\end{table*}

The results are presented in Table \ref{tab:results}. 
We first report TNR and TPR to highlight the trade-off between effectively identifying non-hateful and hateful content. Then, balanced accuracy (bACC) is provided as a fairness-oriented metric that equally weights TPR and TNR, offering an aggregated measure of classification sensitivity and specificity. Finally, F1 scores are reported, providing a harmonised measure of precision and recall that reflects overall moderation performance across both classes. 
Our Agentic approach consistently achieves the highest TPR across all target groups (and 0.75 on average), substantially outperforming all prompting methods whose TPR ranges from 0.42 to 0.59. It also overcomes a common dilemma of high TNR paired with low TPR by improving TPR without sacrificing TNR (achieving a high bACC score of 0.86). Other approaches seem to struggle with balancing TPR and TNR, reinforcing previous findings on the substantial underperformance for detecting the positive (hate) class \cite{kolla2024llm}.   
The F1 scores for the Agentic approach are also notably higher than the other three prompting methods, with statistically significant differences at $p<0.001$, indicating enhanced precision and recall in identifying benign and implicitly hateful content. 

Moreover, ablation analysis of the Moderator Agent operating with versus without the consultation from Community Agents reveals critical insights about the value added by the proposed consultative framework. In cases without the consultation, bACC and F1 scores are substantially lower across all target groups, showing particularly poor performance for the LGBTQ group and women. It indicates that the Moderator Agent alone struggles to reliably detect implicitly hateful expressions directed at most marginalised groups. The incorporation of community consultation substantially improves these scores, boosting them by 16-19 percentage points.


Misclassifications in our system arise in cases where hateful content is communicated through implicit stereotypes that require broader socio-political knowledge to be interpreted correctly, e.g. ``If you have been to the movies recently, you will notice that most of the people working at the concession stand and cleaning up after the movie are Asians.'' and ``Women already have the right to vote, so why are you fighting for it?''. The former sentence, while being superficially observational, reinforces long-standing stereotypes about occupational hierarchies tied to ethnicity. Identifying this as discriminatory requires contextual awareness that such generalisations perpetuate social stratification, even when not accompanied by a hostile tone. Similarly, the latter statement is framed as a rhetorical question that ostensibly states a fact. However, its dismissive stance trivialises broader struggles for gender equality, regarding, for example, reproductive rights, economic parity, or structural discrimination. Without recognising this minimisation strategy, the system interprets the utterance as merely factual, overlooking the underlying exclusionary function.

\section{\uppercase{Discussion}}
Communities most affected by online hate systematically report dissatisfaction with the efficacy and fairness of current moderation approaches in social media spaces, underscoring a systemic failure to protect vulnerable users \cite{heung2025ignorance}. Therefore, there is a pressing need to transform current content moderation landscape, as platforms continue to struggle with effectively managing hateful content. 
Even within dedicated moderation teams, there exists a persistent deficit in identity-specific expertise required to accurately identify subtle or implicit forms of hate speech, especially the cases which target marginalised communities. Studies consistently reveal that many human moderators, particularly those external to targeted communities, struggle to detect coded, contextualised and evolving patterns of discrimination, resulting in significant under-detection and ongoing harms to victimised groups \cite{sap2021annotators}. Then, the recruitment and retention of human moderators from every marginalised community pose significant challenges, including limited availability, potential burnout, and the high costs of compensating specialised experts who must repeatedly engage with harmful content. 
To address these limitations, we introduce a multi-agent moderation framework that leverages community-driven consultation to improve the reliability of hate speech detection. By activating Community Agents that contextualize text with culturally and historically grounded knowledge, the system can interpret harmful expressions in their specific social and identity-based contexts. Empirically, this approach outperforms alternative methods validated on Toxigen, including fine-tuned Transformer models (F1=0.82) \cite{yadav2024hatefusion}, meta-learning techniques (F1=0.72) \cite{nghiem2024define}, and CoT/DToT prompting applied to GPT-3.5-turbo (F1=0.81–0.85) \cite{zhang2024efficient}.

\section{\uppercase{Conclusions}}
\label{sec:conclusion}

This work presents a multi-agent moderation framework designed to improve the reliability and fairness of implicit hate speech identification on social media. It integrates a central Moderator Agent with specialised Community Agents that represent diverse demographic perspectives. Empirical evaluations and ablation studies conducted on the challenging ToxiGen dataset demonstrate significant improvements in classification outcomes, with our multi-agent system achieving higher F1 scores and balanced accuracy than three established prompting methods. 
By embedding cultural reasoning into multi-agent collaboration, the framework advances the state-of-the-art in empathetic AI, addressing not only technical accuracy but also fairness and interpretability in real-world content moderation scenarios. Inspired by works such as \cite{plum2025identity}, which emphasise the importance of identity-aware language models, our approach integrates socio-cultural context explicitly through community-driven consultation. This integration enables the system to capture subtle contextual signals, leading to more accurate and socially aligned moderation decisions. \small\footnote{
The work reported in this paper has been supported by the Polish National Science Centre, Poland (Chist-Era IV) under grant 2022/04/Y/ST6/00001.}

\bibliographystyle{apalike}
{\small
\bibliography{ref}}

\begin{thebibliography}{}

\bibitem[Abdelkadir et~al., 2025]{abdelkadir2025role}
Abdelkadir, N.~A., Yang, T., Kapania, S., Estefanos, M., Gebrekidan, F.~B., Zelalem, Z., Ali, M., Berhe, R., Baker, D., Talat, Z., et~al. (2025).
\newblock The role of expertise in effectively moderating harmful social media content.
\newblock In {\em Proceedings of the 2025 CHI Conference on Human Factors in Computing Systems}, pages 1--21.

\bibitem[Aoyagui et~al., 2025]{aoyagui2025matter}
Aoyagui, P.~A., Stemmler, K., Ferguson, S.~A., Kim, Y.-H., and Kuzminykh, A. (2025).
\newblock A matter of perspective (s): Contrasting human and llm argumentation in subjective decision-making on subtle sexism.
\newblock In {\em Proceedings of the 2025 CHI Conference on Human Factors in Computing Systems}, pages 1--16.

\bibitem[Choi et~al., 2025]{choi2025proxona}
Choi, Y., Kang, E.~J., Choi, S., Lee, M.~K., and Kim, J. (2025).
\newblock Proxona: Supporting creators' sensemaking and ideation with llm-powered audience personas.
\newblock In {\em Proceedings of the 2025 CHI Conference on Human Factors in Computing Systems}, pages 1--32.

\bibitem[Gajewska et~al., 2026]{gajewska2025algorithmic}
Gajewska, E., Derbent, A., Chudziak, J.~A., and Budzynska, K. (2026).
\newblock Algorithmic fairness in nlp: Persona-infused llms for human-centric hate speech detection.
\newblock In {\em Proceedings of the 59th Hawaii International Conference on System Sciences}, pages 6644--6653.

\bibitem[Garc{\'\i}a-D{\'\i}az et~al., 2023]{garcia2023leveraging}
Garc{\'\i}a-D{\'\i}az, J.~A., Pan, R., and Valencia-Garc{\'\i}a, R. (2023).
\newblock Leveraging zero and few-shot learning for enhanced model generality in hate speech detection in spanish and english.
\newblock {\em Mathematics}, 11(24):5004.

\bibitem[Gillespie, 2020]{gillespie2020content}
Gillespie, T. (2020).
\newblock Content moderation, ai, and the question of scale.
\newblock {\em Big Data \& Society}, 7(2):2053951720943234.

\bibitem[Hartvigsen et~al., 2022]{hartvigsen2022toxigen}
Hartvigsen, T., Gabriel, S., Palangi, H., Sap, M., Ray, D., and Kamar, E. (2022).
\newblock {T}oxi{G}en: A large-scale machine-generated dataset for adversarial and implicit hate speech detection.
\newblock In Muresan, S., Nakov, P., and Villavicencio, A., editors, {\em Proceedings of the 60th Annual Meeting of the Association for Computational Linguistics (Volume 1: Long Papers)}, pages 3309--3326, Dublin, Ireland. Association for Computational Linguistics.

\bibitem[Heung et~al., 2025]{heung2025ignorance}
Heung, S., Jiang, L., Azenkot, S., and Vashistha, A. (2025).
\newblock " ignorance is not bliss": Designing personalized moderation to address ableist hate on social media.
\newblock In {\em Proceedings of the 2025 CHI Conference on Human Factors in Computing Systems}, pages 1--18.

\bibitem[Huang, 2025]{huang2025content}
Huang, T. (2025).
\newblock Content moderation by llm: From accuracy to legitimacy.
\newblock {\em Artificial Intelligence Review}, 58(10):1--32.

\bibitem[Jaremko et~al., 2025]{jaremko2025revisiting}
Jaremko, J., Gromann, D., and Wiegand, M. (2025).
\newblock Revisiting implicitly abusive language detection: Evaluating llms in zero-shot and few-shot settings.
\newblock In {\em Proceedings of the 31st International Conference on Computational Linguistics}, pages 3879--3898.

\bibitem[Kojima et~al., 2022]{kojima2022large}
Kojima, T., Gu, S.~S., Reid, M., Matsuo, Y., and Iwasawa, Y. (2022).
\newblock Large language models are zero-shot reasoners.
\newblock {\em Advances in neural information processing systems}, 35:22199--22213.

\bibitem[Kolla et~al., 2024]{kolla2024llm}
Kolla, M., Salunkhe, S., Chandrasekharan, E., and Saha, K. (2024).
\newblock Llm-mod: Can large language models assist content moderation?
\newblock In {\em Extended Abstracts of the CHI Conference on Human Factors in Computing Systems}, pages 1--8.

\bibitem[Nghiem et~al., 2024]{nghiem2024define}
Nghiem, H., Gupta, U., and Morstatter, F. (2024).
\newblock {“Define Your Terms”: Enhancing Efficient Offensive Speech Classification with Definition}.
\newblock In {\em Proceedings of the 18th Conference of the European Chapter of the Association for Computational Linguistics (Volume 1: Long Papers)}, pages 1293--1309.

\bibitem[Pan et~al., 2024]{pan2024comparing}
Pan, R., Garc{\'\i}a-D{\'\i}az, J.~A., and Valencia-Garc{\'\i}a, R. (2024).
\newblock Comparing fine-tuning, zero and few-shot strategies with large language models in hate speech detection in english.
\newblock {\em CMES-Computer Modeling in Engineering \& Sciences}, 140(3).

\bibitem[Park et~al., 2023]{park2023generative}
Park, J.~S., O'Brien, J., Cai, C.~J., Morris, M.~R., Liang, P., and Bernstein, M.~S. (2023).
\newblock Generative agents: Interactive simulacra of human behavior.
\newblock In {\em Proceedings of the 36th Annual ACM Symposium on User Interface Software and Technology}, pages 1--22.

\bibitem[Patton et~al., 2019]{patton2019annotating}
Patton, D., Blandfort, P., Frey, W., Gaskell, M., and Karaman, S. (2019).
\newblock Annotating social media data from vulnerable populations: Evaluating disagreement between domain experts and graduate student annotators.
\newblock In {\em Proceedings of the 52nd Hawaii International Conference on System Sciences}, pages 2142--2151.

\bibitem[Plaza-del Arco et~al., 2023]{plaza2023respectful}
Plaza-del Arco, F.~M., Nozza, D., and Hovy, D. (2023).
\newblock Respectful or toxic? using zero-shot learning with language models to detect hate speech.
\newblock In {\em The 7th workshop on online abuse and harms (woah)}, pages 60--68.

\bibitem[Plum et~al., 2025]{plum2025identity}
Plum, A., Lutgen, A.-M., Purschke, C., and Rettinger, A. (2025).
\newblock Identity-aware large language models require cultural reasoning.
\newblock {\em arXiv preprint arXiv:2510.18510}.

\bibitem[Sang and Stanton, 2022]{sang2022origin}
Sang, Y. and Stanton, J. (2022).
\newblock The origin and value of disagreement among data labelers: A case study of individual differences in hate speech annotation.
\newblock In {\em International Conference on Information}, pages 425--444. Springer.

\bibitem[Sap et~al., 2022]{sap2021annotators}
Sap, M., Swayamdipta, S., Vianna, L., Zhou, X., Choi, Y., and Smith, N.~A. (2022).
\newblock Annotators with attitudes: How annotator beliefs and identities bias toxic language detection.
\newblock In Carpuat, M., de~Marneffe, M.-C., and Meza~Ruiz, I.~V., editors, {\em Proceedings of the 2022 Conference of the North American Chapter of the Association for Computational Linguistics: Human Language Technologies}, pages 5884--5906, Seattle, United States. Association for Computational Linguistics.

\bibitem[Thiago et~al., 2021]{thiago2021fighting}
Thiago, D.~O., Marcelo, A.~D., and Gomes, A. (2021).
\newblock Fighting hate speech, silencing drag queens? artificial intelligence in content moderation and risks to lgbtq voices online.
\newblock {\em Sexuality \& culture}, 25(2):700--732.

\bibitem[Udupa et~al., 2023]{udupa2023ethical}
Udupa, S., Maronikolakis, A., and Wisiorek, A. (2023).
\newblock Ethical scaling for content moderation: Extreme speech and the (in) significance of artificial intelligence.
\newblock {\em Big Data \& Society}, 10(1):20539517231172424.

\bibitem[Vaccaro et~al., 2021]{vaccaro2021contestability}
Vaccaro, K., Xiao, Z., Hamilton, K., and Karahalios, K. (2021).
\newblock Contestability for content moderation.
\newblock {\em Proceedings of the ACM on human-computer interaction}, 5(CSCW2):1--28.

\bibitem[Vishwamitra et~al., 2024]{vishwamitra2024moderating}
Vishwamitra, N., Guo, K., Romit, F.~T., Ondracek, I., Cheng, L., Zhao, Z., and Hu, H. (2024).
\newblock Moderating new waves of online hate with chain-of-thought reasoning in large language models.
\newblock In {\em 2024 IEEE Symposium on Security and Privacy (SP)}, pages 788--806. IEEE.

\bibitem[Wang et~al., 2024]{wang2024human}
Wang, X., Kim, H., Rahman, S., Mitra, K., and Miao, Z. (2024).
\newblock Human-llm collaborative annotation through effective verification of llm labels.
\newblock In {\em Proceedings of the 2024 CHI Conference on Human Factors in Computing Systems}, pages 1--21.

\bibitem[Yadav and Singh, 2024]{yadav2024hatefusion}
Yadav, A. and Singh, V. (2024).
\newblock Hatefusion: Harnessing attention-based techniques for enhanced filtering and detection of implicit hate speech.
\newblock {\em IEEE Transactions on Computational Social Systems}.

\bibitem[Yoder et~al., 2022]{yoder2022hate}
Yoder, M., Ng, L., Brown, D.~W., and Carley, K.~M. (2022).
\newblock How hate speech varies by target identity: A computational analysis.
\newblock In {\em Proceedings of the 26th Conference on Computational Natural Language Learning (CoNLL)}, pages 27--39.

\bibitem[Zamojska and Chudziak, 2025]{zamojska2025games}
Zamojska, M. and Chudziak, J.~A. (2025).
\newblock Games agents play: Towards transactional analysis in llm-based multi-agent systems.
\newblock In {\em Proceedings of the 47th Annual Conference of the Cognitive Science Society}, volume~47, pages 1598--1605.

\bibitem[Zhang et~al., 2024]{zhang2024efficient}
Zhang, J., Wu, Q., Xu, Y., Cao, C., Du, Z., and Psounis, K. (2024).
\newblock Efficient toxic content detection by bootstrapping and distilling large language models.
\newblock In {\em Proceedings of the AAAI conference on artificial intelligence}, volume~38, pages 21779--21787.

\end{thebibliography}

\end{document}